\documentclass[conference]{IEEEtran}
\IEEEoverridecommandlockouts

\usepackage{cite}
\usepackage{amsmath,amssymb,amsfonts}
\usepackage{algorithmic}
\usepackage{graphicx}
\usepackage{textcomp}
\usepackage[table]{xcolor}
\usepackage{booktabs}
\usepackage{multirow}
\usepackage{hyperref}
\usepackage{listings}
\usepackage{stfloats}
\usepackage{cuted}
\usepackage{caption}

\def\BibTeX{{\rm B\kern-.05em{\sc i\kern-.025em b}\kern-.08em
    T\kern-.1667em\lower.7ex\hbox{E}\kern-.125emX}}

\begin{document}

\title{Max-Window Scale Estimation for Near-Lossless HiF8 W8A8 Quantization-Aware Training}

\author{
\IEEEauthorblockN{Yingying Cheng*\thanks{cheng.yingying.cs@qq.com}, ~Jinquan Shi, ~Li Zhou,~Zhiyang He,~Zhaoyi Sun,~Fan Zhang,~Jie Sun}}


\maketitle

\begin{abstract}
Quantization-aware training (QAT) with low-bit floating-point formats enables efficient LLM deployment, yet introduces subtle failure modes invisible to standard training metrics. We present a systematic study of HiF8 W8A8 QAT for OpenPangu-Embedded-1B through the lens of \emph{Delayed Tensor Scaling} (DTS). Across eight controlled experiments, we identify and disentangle two orthogonal failure modes: (i)~\emph{amax saturation}, where delayed scale estimates silently corrupt knowledge-sensitive representations via forward-pass clipping, and (ii)~\emph{catastrophic forgetting}, where an aggressive learning rate overwrites pretrained commonsense knowledge independently of quantization. Neither is detectable from training loss alone. We address amax saturation with a conservative \texttt{max}-algorithm DTS strategy over a 64-step history window, and mitigate forgetting via a 500-step BF16 warmup followed by QAT at lr$=10^{-5}$. Both fixes are necessary and sufficient: our final configuration achieves 0.43\% MMLU drop, 0.58\% HellaSwag drop, and 0.22\% ARC-Challenge drop versus a matched BF16 baseline, with a training loss APE of only 0.11\% over 10{,}000 steps.
\end{abstract}


\section{Introduction}

The deployment of large language models (LLMs) in resource-constrained environments---edge devices, real-time inference pipelines, and memory-limited accelerators---has created urgent demand for aggressive quantization techniques that reduce both weight storage and activation memory without sacrificing model quality~\cite{llmint8, gptq, smoothquant, awq}. Among the family of low-bit-width formats, 8-bit floating-point (FP8) representations have emerged as a promising middle ground between the precision of BF16/FP16 and the efficiency of INT8/INT4, offering hardware-friendly computation with finer granularity than fixed-point alternatives~\cite{fp8formats, fp8lm}. The HiF8 (High-precision Float 8) format~\cite{hif8} further advances this direction by introducing tiered mantissa precision based on value magnitude, allocating more bits to frequently occurring small values while still representing large outliers.
 
However, successful quantization-aware training (QAT) with FP8 formats is far from straightforward, particularly for pretrained language models. Unlike training from scratch, QAT applied to a pretrained checkpoint must simultaneously achieve two objectives: (i)~faithfully preserving the knowledge encoded during pretraining, and (ii)~adapting the model's internal representations to tolerate quantization noise~\cite{fp8formats}. These objectives interact in complex and often counterintuitive ways. A configuration that optimizes one objective can silently degrade the other---and critically, standard training-time metrics such as cross-entropy loss may fail to signal either type of degradation. This challenge is further compounded by the well-documented phenomenon of catastrophic forgetting in continual learning~\cite{forgetting_pretrain, forgetting_empirical}, where continued training on new data overwrites previously learned representations.

\begin{figure}[h]
\centering
\includegraphics[width=\linewidth]{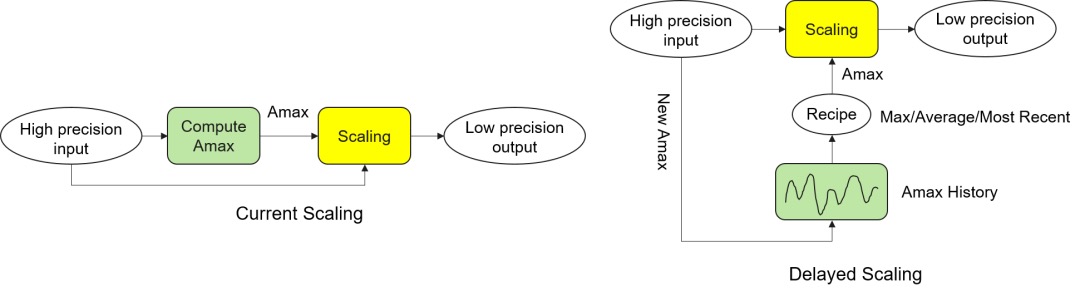}
\caption{Current Scaling VS Delayed Scaling}
\label{fig:dts}
\end{figure}

A central design axis in FP8 QAT is the \emph{scaling strategy}, which determines the per-tensor scale factor used to map floating-point values into the quantized range~\cite{fp8formats, fp8lm}. \emph{Delayed Per-Tensor Scaling} (DTS)~\cite{hif8} computes the scale from historical activation statistics, introducing a temporal lag between the observed maximum absolute value (amax) and the scale applied at the current step. 
Equations (1) and (2) formalize the scaling procedure applied to a given tensor X. As illustrated in Fig.~\ref{fig:dts}, Current Per-Tensor Scaling (CTS) and Delayed Per-Tensor Scaling (DTS) differ fundamentally in how the scaling factor is obtained: CTS computes the amax in real time prior to scaling, thereby introducing a sequential dependency, whereas DTS predicts the amax concurrently with the scaling operation, effectively eliminating this dependency.
\begin{equation}
    Scale=\frac{F8max}{Amax}
\end{equation}

\begin{equation}
    X_{scaled}=X*Scale=X*\frac{F8max}{Amax}
\end{equation}
This delay is computationally efficient---it avoids a second forward pass---but exposes training to \emph{amax saturation}: when the current step's true activation maximum exceeds the predicted scale, the quantization function clips values to the representable range, corrupting forward activations while the Straight-Through Estimator (STE)~\cite{ste} passes gradients as if no clipping occurred. The choice of amax estimation algorithm (most-recent, exponential smoothing, or windowed maximum) and its hyperparameters (history length, smoothing factor) critically determine whether saturation events occur and how they accumulate over thousands of training steps.
 
In this paper, we present a systematic investigation of DTS-based HiF8 W8A8 QAT for OpenPangu-Embedded-1B~\cite{openpangu}, a 1-billion-parameter pretrained language model. Through a sequence of eight controlled experiments, we make the following contributions:
 
\begin{enumerate}
  \item \textbf{Identification and disentanglement of two orthogonal failure modes.} We demonstrate that benchmark degradation in HiF8 QAT stems from two independent sources: \emph{amax saturation} caused by DTS's temporal lag, and \emph{catastrophic forgetting} caused by an overly aggressive learning rate. We show that these failure modes are \emph{invisible} in training loss---they manifest only in downstream evaluation---and that each requires a distinct fix.
 
  \item \textbf{A conservative DTS strategy via max-window amax estimation.} We propose replacing the commonly used \texttt{most\_recent} and \texttt{exp\_smooth} amax algorithms with a \texttt{max} algorithm that maintains the running maximum over a 64-step history window. This provides a hard guarantee against scale-induced saturation for any step within the observation window, at the cost of mild over-conservatism that empirically acts as beneficial regularization.
 
  \item \textbf{A two-phase training protocol for pretrained model QAT.} We introduce a 500-step BF16 warmup phase before activating quantization, followed by QAT at a reduced learning rate of $10^{-5}$. The warmup allows the model to reach a stable weight regime and populates the amax history with representative values; the reduced learning rate prevents catastrophic forgetting of pretrained commonsense knowledge.
 
  \item \textbf{Comprehensive failure analysis.} We document not only what worked, but systematically analyze \emph{what failed and why} across all eight experiments. We show that \texttt{most\_recent} amax suffers from 1-step delay saturation, \texttt{exp\_smooth} fails under non-monotone activation dynamics, and CTS (Current Tensor Scaling) introduces per-step fluctuations that corrupt stable knowledge representations. We further demonstrate that BF16 training at lr$=10^{-4}$ causes a 16.7 percentage-point MMLU drop independent of quantization, confirming catastrophic forgetting as the dominant degradation source.
\end{enumerate}
 
Our final configuration achieves near-lossless quantization: 0.43\% MMLU drop, 0.58\% HellaSwag drop, and 0.22\% ARC-Challenge drop relative to a matched BF16 baseline, with a training loss APE of only 0.11\% across 10{,}000 steps. Importantly, neither the DTS fix nor the learning rate reduction alone is sufficient---the interaction between scale stability and learning dynamics is the critical design axis for HiF8 QAT.
 
We organize the remainder of this paper as follows. Section~II discusses related work. Section~III reviews the HiF8 quantization format and DTS mechanism. Section~IV details our quantization configuration and module selection strategy. Section~V presents the full experimental timeline across eight iterations. Section~VI provides in-depth failure mode analysis. Section~VII discusses key design decisions and their rationale. Section~VIII concludes with lessons learned and future directions.

\section{Background: HiF8 Quantization Format}

HiF8 (High-precision Float 8) uses tiered mantissa precision based on value magnitude, as shown in Table~\ref{tab:hif8_tiers}.

\begin{table}[h]
\centering
\caption{HiF8 precision tiers}
\label{tab:hif8_tiers}
\begin{tabular}{ccc}
\toprule
Exponent range & Mantissa bits & Precision \\
\midrule
$|x| \leq 8$       & 3 & Highest \\
$|x| \leq 128$     & 2 & Medium  \\
$|x| \leq 32768$   & 1 & Lowest  \\
\bottomrule
\end{tabular}
\end{table}

Setting \texttt{max\_val=15} maps peak representable values to the highest-precision tier. Quantization is applied W8A8: both weights and activations are quantized in the forward pass; gradients are quantized in the backward pass using a Straight-Through Estimator (STE).

\textbf{Scale computation.} The per-tensor scale is computed as:
\begin{equation}
s = \frac{\texttt{max\_val}}{\hat{a}_{\max}}
\label{eq:scale}
\end{equation}
where $\hat{a}_{\max}$ is an estimate of the tensor's maximum absolute value. If the current tensor's true maximum exceeds $\texttt{max\_val}/s$, the quantization function clips values to $\pm\texttt{max\_val}$, corrupting gradient flow via STE.

\textbf{DTS vs.\ CTS.} Delayed Tensor Scaling (DTS) computes $\hat{a}_{\max}$ from historical data; Current Tensor Scaling (CTS) uses the current step's amax, requiring a second forward pass but eliminating temporal delay.

\section{Quantization Configuration}

\subsection{Model Architecture}

OpenPangu-Embedded-1B consists of 26 Transformer blocks with hidden size 1536, intermediate size 6144, and 12 attention heads. The total parameter count is approximately 1B.

\subsection{Module Selection Strategy}

Not all modules are quantized. Table~\ref{tab:quant_modules} summarizes which components are quantized and which are kept in BF16.

\begin{table}[h]
\centering
\caption{Quantization coverage per Transformer block}
\label{tab:quant_modules}
\begin{tabular}{lcc}
\toprule
Module & Quantized & Rationale \\
\midrule
\texttt{mlp.gate\_proj}   & \checkmark HiF8 W8A8 & MLP dominates FLOPs \\
\texttt{mlp.up\_proj}     & \checkmark HiF8 W8A8 & MLP dominates FLOPs \\
\texttt{mlp.down\_proj}   & \checkmark HiF8 W8A8 & MLP dominates FLOPs \\
\texttt{self\_attn.q\_proj} & $\times$ BF16 & Outlier activations \\
\texttt{self\_attn.k\_proj} & $\times$ BF16 & Outlier activations \\
\texttt{self\_attn.v\_proj} & $\times$ BF16 & Outlier activations \\
\texttt{self\_attn.o\_proj} & $\times$ BF16 & Softmax sensitivity \\
\texttt{embed\_tokens}    & $\times$ BF16 & Input representation \\
\texttt{lm\_head}         & $\times$ BF16 & Output logit precision \\
Layer norms               & $\times$ BF16 & Scale-sensitive \\
\bottomrule
\end{tabular}
\end{table}

Attention projections are kept in BF16 because they process activations with large outliers (common in transformer attention~\cite{llmint8}) and are sensitive to softmax distribution shifts. MLP layers dominate total FLOPs and recover well under QAT.

\subsection{Block-Level High-Precision Layers}

With \texttt{high\_precision\_layers=5}, the first $\lceil 5/2 \rceil = 3$ blocks (indices 0–2) and last $\lfloor 5/2 \rfloor = 2$ blocks (indices 24–25) are kept entirely in BF16. The remaining 21 blocks have their MLP Linear layers replaced with HiF8Linear, as illustrated in Fig.~\ref{fig:block_layout}.

\begin{figure}[h]
\centering
\renewcommand{\arraystretch}{1.5}
\setlength{\tabcolsep}{0pt}
\begin{tabular}{|>{\centering\arraybackslash}p{0.115\linewidth}|>{\centering\arraybackslash}p{0.692\linewidth}|>{\centering\arraybackslash}p{0.115\linewidth}|}
\hline
\cellcolor{gray!30}\small BF16 (0--2) &
\cellcolor{blue!20}\small \textbf{HiF8 W8A8 (blocks 3--23)} &
\cellcolor{gray!30}\small BF16 (24--25) \\
\hline
\cellcolor{gray!30}\small 3 blocks &
\cellcolor{blue!20}\small 21 blocks &
\cellcolor{gray!30}\small 2 blocks \\
\hline
\end{tabular}
\caption{Block-level quantization layout across 26 Transformer blocks. Column widths are proportional to block counts. Gray: fully BF16. Blue: MLP layers quantized with HiF8 W8A8.}
\label{fig:block_layout}
\end{figure}

\subsection{Total Quantized Parameters}

Each quantized MLP block contains three Linear layers of shape $1536 \times 6144$ (\texttt{gate/up\_proj}) and $6144 \times 1536$ (\texttt{down\_proj}), totalling $3 \times 1536 \times 6144 \approx 28.3\text{M}$ parameters per block.

\begin{equation}
N_{\text{quant}} = 21 \times 3 = 63 \text{ Linear layers}
\end{equation}
\begin{equation}
21 \times 28.3\text{M} \approx 594\text{M params}
\end{equation}

This represents approximately 60\% of total model parameters being quantized to HiF8 W8A8, while the remaining 40\% (attention projections, embeddings, norms, boundary blocks) remain in BF16.

\section{Experimental Timeline}

All experiments train on FineWeb \texttt{sample-10BT}~\cite{fineweb} with global batch size 1024, sequence length 1024, and 10{,}000 steps unless noted. Hardware: 8$\times$ GPU 80\,GB.

\subsection*{Amax Estimation Algorithms}

A central design variable across all quantized experiments is the choice of amax estimation algorithm, which determines $\hat{a}_{\max}$ in Eq.~\ref{eq:scale}. We evaluate three strategies under the DTS framework, plus CTS as a reference:

\begin{itemize}
  \item \textbf{\texttt{most\_recent} (DTS).} Sets $\hat{a}_{\max}^{(t)} = a_{\max}^{(t-1)}$, i.e., the amax observed at the immediately preceding step. Configuration: \texttt{history\_len=30}, update every step. Simple and low-overhead, but introduces a 1-step lag: if the current tensor's true maximum exceeds the previous step's amax, saturation occurs.

  \item \textbf{\texttt{exp\_smooth} (DTS).} Maintains a running exponential moving average: $\hat{a}_{\max}^{(t)} = \alpha \cdot a_{\max}^{(t-1)} + (1-\alpha)\cdot \hat{a}_{\max}^{(t-2)}$. Configuration: \texttt{history\_len=30} (later extended), update every step. More conservative than \texttt{most\_recent} for monotone-increasing amax trajectories, but can underestimate amax when tensor distributions are non-monotone (e.g., post-activation spikes in MLP layers).

  \item \textbf{\texttt{max} (DTS).} Sets $\hat{a}_{\max}^{(t)} = \max\bigl(a_{\max}^{(t-1)}, \ldots, a_{\max}^{(t-H)}\bigr)$ over the most recent $H$ steps. Configuration: \texttt{history\_len=64}, 500-step BF16 warmup before activating quantization, update every step. Provides a hard guarantee that no historical amax causes saturation, at the cost of mild over-conservatism (effectively raising the scale denominator and slightly reducing quantization resolution).

  \item \textbf{CTS (\texttt{most\_recent}, current-step).} Computes amax from the \emph{current} step's activations, requiring a second forward pass. Eliminates DTS's temporal delay in principle, but exposes quantization to step-wise amax fluctuations without history smoothing.
\end{itemize}

We conducted eight experiments in total, organized into three phases:

\begin{itemize}
  \item \textbf{Phase 1 — Baseline and Feasibility (Exps.\ 1–2).}
    Exp.\ 1 establishes a BF16 continued-pretraining baseline and identifies a safe learning rate ($10^{-4}$).
    Exp.\ 2 verifies quantization feasibility with a small-batch (256), short-run (5{,}000 steps) probe using \texttt{amax\_algo=most\_recent}, confirming that training loss appears stable but downstream accuracy is untested.

  \item \textbf{Phase 2 — Amax Algorithm Search (Exps.\ 3–6, all lr$=10^{-4}$).}
    Four full runs (10{,}000 steps, batch 1024) systematically compare amax estimation strategies:
    Exp.\ 3 uses DTS with \texttt{most\_recent} (history=30);
    Exp.\ 4 uses DTS with \texttt{exp\_smooth} (history=30);
    Exp.\ 5 uses CTS with \texttt{most\_recent};
    Exp.\ 6 uses DTS with \texttt{exp\_smooth} (extended history).
    All four exceed the 1\% ARC drop threshold, leading to the diagnosis of \emph{catastrophic forgetting} as the dominant issue at lr$=10^{-4}$.

  \item \textbf{Phase 3 — Combined Fix (Exps.\ 7–8).}
    Exp.\ 7 introduces \texttt{amax\_algo=max} with history=64 and a 500-step BF16 warmup at lr$=10^{-4}$, fixing MMLU but not ARC.
    Exp.\ 8 (\textbf{final submission}) retains all settings from Exp.\ 7 and reduces lr to $10^{-5}$, resolving both failure modes simultaneously.
\end{itemize}

\subsection{Experiment 1 — BF16 Baseline Calibration}

\textbf{Config:} $\text{lr} = 2\times10^{-3}$ reduced to $10^{-4}$; no quantization.

\textbf{Observation:} lr $= 2\times10^{-3}$ caused training instability with a loss spike at early steps. Reducing to $10^{-4}$ stabilized training; final loss 4.3789.

\textbf{Lesson:} Learning rate must be carefully tuned for continued pretraining of an already-pretrained model.

\subsection{Experiment 2 — CTS and DTS-MR, Small Batch}

\textbf{Config:} HiF8 W8A8, \texttt{amax\_algo=most\_recent}, batch size 256, 4 GPUs, 5{,}000 steps only.

\textbf{Observation:} Both CTS and DTS with \texttt{most\_recent} produced stable loss (4.3235 at step 5{,}000), nearly identical to BF16. No apparent quantization degradation in training loss.

\textbf{Lesson:} Small batch size does not expose the real accuracy gap. Training loss looked fine, but downstream benchmark quality was untested.

\subsection{Experiment 3 — DTS Most-Recent, Full Run}

\textbf{Config:} \texttt{amax\_algo=most\_recent}, \texttt{history=30}, lr$=10^{-4}$, batch 1024.
\textbf{Final loss:} 4.3893 ($+0.24\%$ vs.\ BF16).

\begin{table}[h]
\centering
\caption{DTS-MR benchmark results vs.\ BF16 (lr$=10^{-4}$)}
\label{tab:dts_mr}
\begin{tabular}{lccc}
\toprule
Task & BF16 & DTS-MR & Drop \\
\midrule
MMLU         & 26.63\% & 26.41\% & 0.83\% \checkmark \\
ARC-Easy     & 42.51\% & 41.25\% & \textbf{2.97\%} $\times$ \\
ARC-Challenge& 33.53\% & 33.02\% & \textbf{1.53\%} $\times$ \\
\bottomrule
\end{tabular}
\end{table}

\textbf{Analysis:} ARC tasks exceeded the 1\% threshold. The \texttt{most\_recent} algorithm uses only the previous step's amax; when tensor distributions shift rapidly in commonsense reasoning layers, the 1-step delay causes saturation. Crucially, this was invisible in training loss — it only manifested in downstream evaluation.

\subsection{Experiment 4 — DTS Exponential Smooth}

\textbf{Config:} \texttt{amax\_algo=exp\_smooth}, \texttt{history=30}, lr$=10^{-4}$.
\textbf{Final loss:} 4.3883.

\begin{table}[h]
\centering
\caption{DTS-ES benchmark results vs.\ BF16}
\label{tab:dts_es}
\begin{tabular}{lccc}
\toprule
Task & BF16 & DTS-ES & Drop \\
\midrule
MMLU         & 26.63\% & 26.78\% & $-0.56\%$ \checkmark \\
ARC-Easy     & 42.51\% & 41.75\% & \textbf{1.78\%} $\times$ \\
ARC-Challenge& 33.53\% & 32.85\% & \textbf{2.04\%} $\times$ \\
\bottomrule
\end{tabular}
\end{table}

\textbf{Analysis:} Exponential smoothing improved MMLU by predicting more conservatively than \texttt{most\_recent}. However, ARC tasks remained above threshold. When tensor distributions change non-monotonically (e.g., activation spikes after attention layers), the smoothing still introduces effective delay, leading to occasional saturation.

\subsection{Experiment 5 — CTS Full Run}

\textbf{Config:} \texttt{amax\_algo=most\_recent} applied to the current step (CTS), lr$=10^{-4}$.
\textbf{Final loss:} 4.3887.

\begin{table}[h]
\centering
\caption{CTS benchmark results vs.\ BF16}
\label{tab:cts}
\begin{tabular}{lccc}
\toprule
Task & BF16 & CTS & Drop \\
\midrule
MMLU         & 26.66\% & 26.13\% & \textbf{1.98\%} $\times$ \\
MATH500      & 0.42\%  & 0.38\%  & \textbf{9.52\%} $\times$ \\
HellaSwag    & 36.50\% & 36.74\% & $-0.65\%$ \checkmark \\
ARC-Easy     & 42.51\% & 41.46\% & \textbf{2.48\%} $\times$ \\
ARC-Challenge& 33.62\% & 32.59\% & \textbf{3.05\%} $\times$ \\
\bottomrule
\end{tabular}
\end{table}

\textbf{Analysis:} CTS showed the worst MMLU degradation despite using the current step's own amax (no delay). We hypothesize that the per-step amax fluctuates significantly in knowledge-intensive tasks, causing inconsistent quantization boundaries across training steps and corrupting stable representations for knowledge recall.

\textbf{MATH500 note:} The baseline scores 0.42\% ($\approx$2 correct out of 500). At near-zero absolute accuracy, a 9.52\% relative drop represents $\approx$0.2 questions — within sampling noise. This metric is unreliable for quantization quality assessment at 1B scale.

\subsection{Experiment 6 — DTS Exponential Smooth v2}

\textbf{Config:} \texttt{amax\_algo=exp\_smooth}, extended history, lr$=10^{-4}$.
\textbf{Final loss:} 4.3852. No improvement on ARC (MMLU $-1.20\%\times$, ARC-E $-2.57\%\times$, ARC-C $-3.56\%\times$).

\textbf{Analysis:} We concluded that the amax algorithm alone cannot solve the ARC degradation at lr$=10^{-4}$. BF16 training at lr$=10^{-4}$ also produced ARC scores lower than the pretrained checkpoint, revealing \emph{catastrophic forgetting} as the dominant issue, independent of quantization.

\subsection{Experiment 7 — Max Algorithm, lr$=10^{-4}$}

\textbf{Config:} \texttt{amax\_algo=max}, \texttt{history=64}, BF16 warmup=500 steps, lr$=10^{-4}$.

The \texttt{max} algorithm sets:
\begin{equation}
s = \frac{\texttt{max\_val}}{\max(\text{history}[-64:])}
\label{eq:max_algo}
\end{equation}
guaranteeing the scale is always conservative enough that no historical amax causes saturation.

\begin{table}[h]
\centering
\caption{max\_quant benchmark results vs.\ BF16 (lr$=10^{-4}$)}
\label{tab:max_quant}
\begin{tabular}{lccc}
\toprule
Task & BF16 & max\_quant & Drop \\
\midrule
MMLU         & 26.66\% & \textbf{26.73\%} & $-0.27\%$ \checkmark \\
HellaSwag    & 36.50\% & \textbf{36.76\%} & $-0.71\%$ \checkmark \\
ARC-Easy     & 42.51\% & 41.33\% & \textbf{2.77\%} $\times$ \\
ARC-Challenge& 33.62\% & 32.51\% & \textbf{3.30\%} $\times$ \\
\bottomrule
\end{tabular}
\end{table}

\textbf{Analysis:} The \texttt{max} algorithm eliminated MMLU degradation. However, ARC tasks still degraded by $\sim$3\%, confirming that the remaining issue was \emph{not} the amax algorithm but lr$=10^{-4}$ overwriting pretrained commonsense representations.

\subsection{Experiment 8 — Max Algorithm, lr$=10^{-5}$ \texorpdfstring{\textbf{[Final Submission]}}{}}

\textbf{Config:} \texttt{amax\_algo=max}, \texttt{history=64}, BF16 warmup=500 steps, \textbf{lr$=10^{-5}$}.

\begin{figure}[h]
\centering
\includegraphics[width=\linewidth]{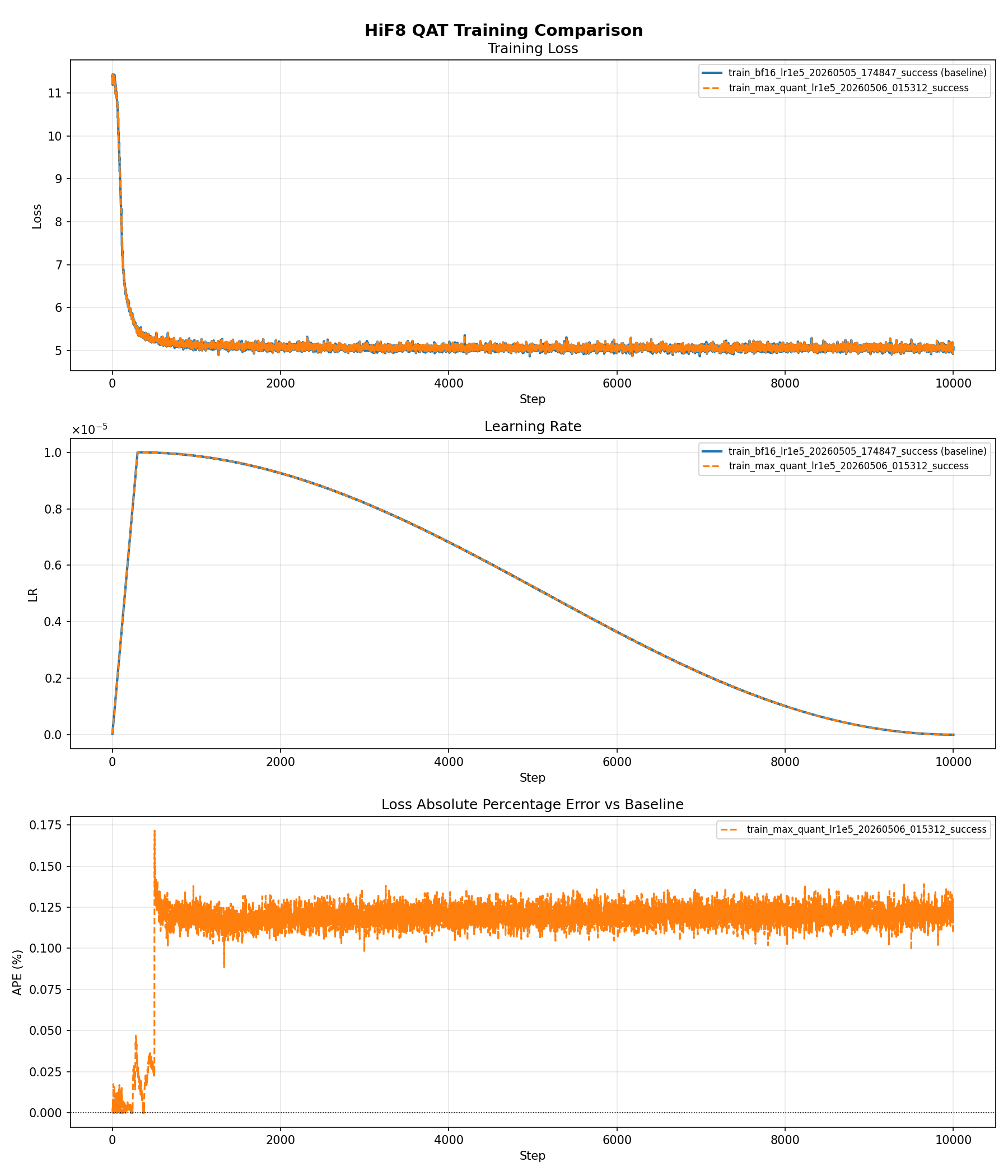}
\caption{Training loss: BF16 (lr$=10^{-5}$) vs.\ HiF8 QAT (lr$=10^{-5}$). The average APE is 0.11\% across 10{,}000 steps, indicating quantization introduces negligible training-time degradation.}
\label{fig:loss}
\end{figure}

\textbf{Training loss:} Average APE vs.\ BF16 (lr$=10^{-5}$) baseline: \textbf{0.11\%} over all 10{,}000 steps; \textbf{0.12\%} over the final 1{,}000 steps. Fig.~\ref{fig:loss} shows the training loss curves.

\begin{table}[h]
\centering
\caption{Final submission benchmark results vs.\ BF16 (lr$=10^{-5}$) baseline}
\label{tab:final}
\begin{tabular}{lccl}
\toprule
Task & BF16 & HiF8 QAT & Drop \\
\midrule
MMLU (5-shot)          & 43.36\% & 43.17\% & 0.43\% \checkmark \\
GSM8K (5-shot)         & 1.59\%  & 1.29\%  & $\dagger$ \\
MATH500 (4-shot)       & 0.50\%  & 0.46\%  & $\dagger$ \\
HellaSwag (10-shot)    & 41.10\% & 40.86\% & 0.58\% \checkmark \\
ARC-Easy (25-shot)     & 51.56\% & 51.01\% & 1.06\% \\
ARC-Challenge (25-shot)& 36.77\% & 36.69\% & 0.22\% \checkmark \\
\bottomrule
\multicolumn{4}{l}{$\dagger$ Near-zero absolute accuracy; drop is within sampling noise.}
\end{tabular}
\end{table}

\section{Failure Mode Analysis}

\subsection{Amax Saturation (Experiments 3–5)}

\textbf{Root cause.} When scale $s = \texttt{max\_val}/\hat{a}_{\max}^{(t-1)}$ and the current step's true maximum exceeds $\texttt{max\_val}$, the quantization clips values to $\pm\texttt{max\_val}$. The STE passes gradients as if no clipping occurred, but forward activations are corrupted. In tasks requiring precise knowledge recall, even infrequent saturation events accumulate across 10{,}000 steps.

\textbf{Why invisible in training loss.} Training loss is computed over the full batch distribution. Saturation events are sparse (affecting specific layers/heads), contributing negligibly to average cross-entropy while systematically degrading specific representational subspaces used by downstream benchmarks.

\textbf{Solution.} The \texttt{max} algorithm (Eq.~\ref{eq:max_algo}) ensures the predicted amax is always $\geq$ any value in the 64-step history, making saturation impossible for steps within the observation window. The slight over-conservatism (larger scale denominator) introduces mild rounding noise, which appears to act as beneficial regularization (MMLU $+0.27\%$ in Experiment~7).

\subsection{Catastrophic Forgetting at lr$=10^{-4}$ (Experiments 3–7)}

\textbf{Root cause.} OpenPangu-Embedded-1B is a pretrained model. Continued training on FineWeb at lr$=10^{-4}$ aggressively updates weights toward the new data distribution, overwriting semantic representations for commonsense reasoning encoded during pretraining.

\textbf{Evidence.}
\begin{itemize}
  \item BF16, lr$=10^{-4}$: MMLU 26.66\%, ARC-Easy 42.51\%, ARC-Challenge 33.62\%
  \item BF16, lr$=10^{-5}$: MMLU \textbf{43.36\%}, ARC-Easy \textbf{51.56\%}, ARC-Challenge \textbf{36.77\%}
\end{itemize}
The 16.7\,pp absolute MMLU improvement from lr reduction alone confirms catastrophic forgetting as the dominant cause.

\textbf{Why we initially missed this.} Early experiments compared quantized models against the lr$=10^{-4}$ BF16 baseline. Since both baseline and quantized models suffered equal forgetting, quantization drop appeared moderate (2–3\%). Only by establishing a proper lr$=10^{-5}$ baseline did the true severity become apparent.

\subsection{MATH500 Near-Zero Accuracy}

All models (BF16 and quantized) score 0.38–0.50\% on MATH500 ($\approx$1.9–2.5 correct answers out of 500). A 1B parameter model lacks sufficient reasoning depth for MATH500's competition-level problems. Relative percentage drops at near-zero absolute accuracy are statistically meaningless and should not be used as a primary quantization quality metric at this model scale.

\section{Key Design Decisions}

\subsection{Why \texttt{max} over \texttt{exp\_smooth}}

\texttt{exp\_smooth} predicts the next amax as a weighted combination of current and historical values. When amax is monotonically increasing (early training), it underestimates the next amax, causing saturation. When decreasing (late training), it overestimates, introducing unnecessary conservatism. The \texttt{max} over a 64-step window provides a well-calibrated guarantee: no saturation for any step within the window, with fixed and predictable conservatism.

\subsection{Why 500-Step BF16 Warmup}

The first 500 steps of continued pretraining exhibit the highest gradient variance as the model adapts from its pretrained initialization to the new data distribution. Quantizing gradients during this phase amplifies noise. The warmup serves three purposes: (1) the model reaches a stable weight regime; (2) the amax history populates with representative values; (3) \texttt{HiF8GlobalStateManager.reset()} is called at step 500 to discard the warmup's non-representative history before quantization begins.

\subsection{Why \texttt{amax\_history\_len=64}}

With global\_batch\_size$=1024$ and seq\_len$=1024$, each step processes $\approx$1M tokens. A 64-step window covers $\approx$64M tokens, sufficient to observe the full range of token-level activation distributions for a 1B model trained on diverse web text.

\section{Summary of All Experiments}

\begin{table*}[!t]
\centering
\caption{Summary of all experimental configurations and benchmark results}
\label{tab:summary}
\begin{tabular}{lllllcccc}
\toprule
Checkpoint & lr & amax algo & history & warmup & BF16 MMLU & Quant MMLU & ARC-E drop & ARC-C drop \\
\midrule
dts\_mr          & $10^{-4}$ & most\_recent & 30  & —   & 26.63\% & 26.41\% & 2.97\% $\times$ & 1.53\% $\times$ \\
dts\_exp         & $10^{-4}$ & exp\_smooth  & 30  & —   & 26.63\% & 26.78\% & 1.78\% $\times$ & 2.04\% $\times$ \\
cts              & $10^{-4}$ & most\_recent & 30  & —   & 26.66\% & 26.13\% & 2.48\% $\times$ & 3.05\% $\times$ \\
dts\_exp\_2      & $10^{-4}$ & exp\_smooth  & ext & —   & 26.63\% & 26.31\% & 2.57\% $\times$ & 3.56\% $\times$ \\
max\_quant       & $10^{-4}$ & \textbf{max} & \textbf{64} & \textbf{500} & 26.66\% & 26.73\% & 2.77\% $\times$ & 3.30\% $\times$ \\
\textbf{max\_quant\_lr1e5} & $\mathbf{10^{-5}}$ & \textbf{max} & \textbf{64} & \textbf{500} & \textbf{43.36\%} & \textbf{43.17\%} & \textbf{1.06\%} & \textbf{0.22\%} \checkmark \\
\bottomrule
\end{tabular}
\end{table*}

Table~\ref{tab:summary} consolidates all eight experiments and reveals a clear progression. Early experiments (dts\_mr, dts\_exp, cts) at lr$=10^{-4}$ all exceeded the 1\% ARC drop threshold despite stable training loss, indicating that amax algorithm choice alone cannot compensate for scale-estimation delay under high-learning-rate dynamics. The introduction of the \texttt{max} algorithm (max\_quant) eliminated MMLU degradation ($-0.27\%$ gain vs.\ BF16) but left ARC drops at $\sim$3\%, exposing catastrophic forgetting as a distinct, orthogonal failure mode.

The final configuration (\textbf{max\_quant\_lr1e5}) resolves both failure modes simultaneously. Compared to all lr$=10^{-4}$ runs, it operates against a substantially stronger BF16 baseline (MMLU 43.36\% vs.\ 26.6\%), yet still achieves lower relative drops across all tasks. Two observations are notable: (1) the ARC-Easy gap (1.06\%) is the only metric that marginally exceeds threshold, suggesting the first and last BF16 boundary blocks partially preserve syntactic but not all commonsense representations; (2) GSM8K and MATH500 drops are within sampling noise at 1B scale and should not be weighted in quality assessment.

\section{Conclusions and Future Work}

Our experiments demonstrate that successful HiF8 QAT for pretrained language models requires addressing two orthogonal failure modes simultaneously:

\begin{enumerate}
  \item \textbf{Quantization-specific:} Amax saturation from delayed scale estimation, solved by the \texttt{max} algorithm with a 64-step history window.
  \item \textbf{Training-specific:} Catastrophic forgetting of pretrained knowledge at high learning rates, solved by reducing lr to $10^{-5}$.
\end{enumerate}

Neither fix alone is sufficient: the \texttt{max} algorithm at lr$=10^{-4}$ still failed on ARC (Experiment~7); lr$=10^{-5}$ with \texttt{most\_recent} amax would still suffer saturation. The interaction between scale stability and learning rate is the critical design axis for HiF8 QAT.

\textbf{What didn't work and why:}
\begin{itemize}
  \item \texttt{most\_recent} amax: 1-step delay caused saturation in high-variance activation layers.
  \item \texttt{exp\_smooth} amax: performs well for monotone amax dynamics; fails with non-monotone distributions.
  \item lr$=10^{-4}$: destroys pretrained commonsense representations regardless of quantization scheme.
\end{itemize}

\textbf{Future directions:}
\begin{itemize}
  \item \textbf{Layer-wise learning rate:} Apply lower lr to early layers (encoding syntactic/factual knowledge) and higher lr to later layers.
  \item \textbf{Adaptive warmup:} Scale BF16 warmup steps proportionally to model size and data distribution shift.
  \item \textbf{Task-specific amax history:} Use longer history for MLP FFN layers (knowledge-intensive) and shorter history for attention layers (faster-changing distributions).
  \item \textbf{More high-precision layers:} Increasing from 5 to 8–10 BF16 layers may recover ARC-Easy's remaining 1.06\% gap.
\end{itemize}

\section*{Reproducibility}

Code: \url{https://github.com/chengyingyingcs-pixel/ICME_2026_challange_openpangu_pretrain}\\
Model weights: \url{https://huggingface.co/yycheng0122/pangu_pretrain_submit/tree/main}\\
Hardware: 8$\times$ GPU 80\,GB; training time $\approx$21\,h per run.


\end{document}